\documentclass[12pt]{article}
\usepackage{graphicx}
\usepackage{multirow}
\usepackage{booktabs}
\usepackage{amsmath}
\usepackage[a4paper, total={6in, 9in}]{geometry}
\usepackage[font=footnotesize]{caption}
\usepackage{hyperref}
\usepackage{lineno}
\usepackage{makecell}
\usepackage[ruled]{algorithm2e}
\usepackage{algpseudocode}
\usepackage{xcolor}
\usepackage{soul} 
\begin{document}
\title{Super Co-alignment of Human and AI for Sustainable Symbiotic Society}

\author
{Yi Zeng$^{1,2,3,4,7,*}$, Feifei Zhao$^{1,2,3,7}$, Yuwei Wang$^{1,2,3,7}$, Enmeng Lu$^{1,2,3,7}$,\\ Yaodong Yang$^{1,2,6}$, Lei Wang$^{1,2,5}$, Chao Liu$^{1,8}$, Yitao Liang$^{1,6}$, \\Dongcheng Zhao$^{1,2,3,7}$, Bing Han$^{3,4}$,  Haibo Tong$^{3,4}$, Yao Liang$^{3,4}$,\\ Dongqi Liang$^{3,4}$,  Kang Sun$^{2,7}$, Boyuan Chen$^{6}$,  Jinyu Fan$^{1,2,3,7}$ \\
\\
\normalsize{$^{1}$Beijing Key Laboratory of Safe AI and Superalignment, China.}\\
\normalsize{$^{2}$Beijing Institute of AI Safety and Governance, China.}\\
\normalsize{$^{3}$Brain-inspired Cognitive AI Lab, Institute of Automation, }\\
\normalsize{Chinese Academy of Sciences, China.}\\
\normalsize{$^{4}$University of Chinese Academy of Sciences, China.}\\
\normalsize{$^{5}$Wenge Technology Co., Ltd.}\\
\normalsize{$^{6}$ Institute for Artificial Intelligence, Peking University, China.}\\
\normalsize{$^{7}$ Long-term AI, China.}\\
\normalsize{$^{8}$ State Key Laboratory of Cognitive Neuroscience and Learning, }\\
\normalsize{Beijing Normal University, China.}\\
\normalsize{$^{*}$Corresponding author: yi.zeng@ia.ac.cn}\\
}
\date{}
\maketitle


\begin{abstract}
As Artificial Intelligence (AI) advances toward Artificial General Intelligence (AGI) and eventually Artificial Superintelligence (ASI), it may potentially surpass human control, deviate from human values, and even lead to irreversible catastrophic consequences in extreme cases. This looming risk underscores the critical importance of the "superalignment" problem - ensuring that AI systems which are much smarter than humans, remain aligned with human (compatible) intentions and values. While current scalable oversight and weak-to-strong generalization methods demonstrate certain applicability, they exhibit fundamental flaws in addressing the superalignment paradigm - notably, the unidirectional imposition of human values cannot accommodate superintelligence's autonomy or ensure AGI/ASI's stable learning. We contend that the values for sustainable symbiotic society should be co-shaped by humans and living AI together, achieving "Super Co-alignment." Guided by this vision, we propose a concrete framework that integrates external oversight and intrinsic proactive alignment. External oversight superalignment should be grounded in human-centered ultimate decision, supplemented by interpretable automated evaluation and correction, to achieve continuous alignment with humanity's evolving values. Intrinsic proactive superalignment is rooted in a profound understanding of the Self, others, and society, integrating self-awareness, self-reflection, and empathy to spontaneously infer human intentions, distinguishing good from evil and proactively prioritizing human well-being. The integration of externally-driven oversight with intrinsically-driven proactive alignment will co-shape symbiotic values and rules through iterative human-ASI co-alignment, paving the way for achieving safe and beneficial AGI and ASI for good, for human, and for a symbiotic ecology.
\end{abstract}

\section*{Keywords}
Super Co-alignment, Human-AI Co-alignment, External Oversight Superalignment, Intrinsic Proactive Superalignment, Sustainable Symbiotic Society

\section{Introduction}
With the breakthrough advancement of Artificial Intelligence (AI), the emergence of large language models (LLMs)~\cite{xi2025rise,zhao2023survey} has achieved human-level and even superhuman performance on multiple benchmarks. This technological leap has directly driven academic and corporate exploration into the theory of Artificial General Intelligence (AGI)~\cite{goertzel2014artificial} and even Artificial Superintelligence (ASI)~\cite{pohl2015artificial}. ASI is defined as ``any intellect that greatly exceeds the cognitive performance of humans in virtually all domains of interest''~
\cite{nick2014superintelligence}. While we promisingly advance the development of increasingly powerful and autonomous general-purpose AI systems, growing awareness of their potential ethical and safety risks has also emerged, such as malicious misuse, loss of control, power-seeking, and strategic deception~\cite{hendrycks2023overview,bengio2024managing}. Indeed, current LLMs have already demonstrated instances of alignment faking~\cite{greenblatt2024alignment}, deception~\cite{park2024ai}, and sycophancy~\cite{sharma2023towards}. When extended to superintelligence, without proper arrangement, it is foreseeable that it may surpass the boundaries of human governance, violate human values, and potentially cause irreversible, uncontrollable, and catastrophic consequences~\cite{russell2019human, nick2014superintelligence}.

Despite experts warning about AI's existential risks~\cite{safeai2023risk}, the development of AI alignment, as well as AI governance framework and ethical safety constraints, still struggles to keep pace with the transformative advancements and rapid iterations of the technology. Superintelligence exceeding human cognitive capacity would possess recursive self-improvement capabilities, achieving exponential advancement rates that would beyond human capacity to monitor or control~\cite{russell2009ethics}. This compels us to proactively address the question, "How do we ensure that AI systems much smarter than humans follow human intentions?" This is OpenAI's definition of superalignment~\cite{Introducing2023}, the key challenge of superalignment is that ASI will far exceed human oversight capabilities, making direct human supervision infeasible. As a result, traditional alignment approaches like Reinforcement Learning from Human Feedback (RLHF)~\cite{bai2022training,ouyang2022training} will fail when confronted with superintelligence more intelligent than humans, as they cannot provide sufficiently high-quality oversight signals to supervise and improve the system.

Current proposed feasible approaches for superalignment such as scalable oversight~\cite{amodei2016concrete,christiano2018supervising} and weak-to-strong generalization~\cite{burns2023weak,taoYourWeakLLM2024a}, aiming to develop scalable high-quality supervision signals, utilize weaker AI systems to guide or supervise stronger AI, ensuring alignment with human values and intentions. Since superintelligent AI systems do not exist yet, researchers have designed experiments to show that scalable oversight is feasible on existing LLMs~\cite{bowman2022measuring}, and that using a "weak-to-strong" approach (supervised GPT-4 with a GPT-2-level model) can meaningfully recover much of GPT-4’s capabilities~\cite{burns2023weak}. By leveraging a lightweight model to learn correctional residuals, Aligner provides model-agnostic guidance, enabling iterative enhancement of large-scale upstream models~\cite{ji2024aligner}. Besides, research on the weak-to-strong generalization approach has progressively extended to enhancing weak-to-strong generalization and alignment capabilities through methods such as benign overfitting~\cite{wuProvableWeaktostrongGeneralization2025}, data-centric lens~\cite{shinWeaktostrongGeneralizationDatacentric2025}, transfer learning framework~\cite{somerstepTransferLearningFramework2025}, weak-to-strong preference optimization~\cite{zhuWeaktostrongPreferenceOptimization2025}, and multi-agent contrastive preference optimization~\cite{lyuMACPOWeaktostrongAlignment2025}. 
However, the weak-to-strong generalization framework presents risks of advanced models developing deceptive behaviors and oversight evasion that remain undetectable to their less capable evaluators~\cite{yangSuperficialalignmentStrongModels2024}. Furthermore, stronger models are still not equivalent to AGI/ASI, and weak-to-strong generalization approaches may fail when applied to genuine AGI/ASI systems, as such systems could exhibit resistance behaviors and lack well-defined motivations to sustain their "learner" roles.

Previously, several scalable oversight techniques have been proposed, including Iterated Distillation and Amplification (IDA)~\cite{Ajeya2018} and Recursive Reward Modeling (RRM)~\cite{Leike2018scalable}, aiming to amplify the scalability of human supervision signals through interactive iterations and subtask decomposition. Faced with superintelligence that surpasses human cognition, traditional human assessment and supervision become untenable and expensive. Recent scalable oversight approaches, such as Reinforcement Learning from AI Feedback (RLAIF)~\cite{lee2023rlaif}, leverages AI-generated feedback to replace human feedback, enabling more precise oversight of AI outcomes with far fewer human labels. Cooperative Inverse Reinforcement Learning (CIRL)~\cite{Hadfield2016cooperative,hadfield2017inverse} and assistance games~\cite{laidlaw2024scalably,laidlaw2025assistancezero} feature AI systems maintaining uncertainty about the reward function, which drives them to actively infer humans' true reward functions through human-AI cooperative interaction. This cooperative and  partial-information game approach achieves alignment with human values. Some approaches employ an additional model to generate red-team testing~\cite{perez2022red}, use external tools for evaluation and feedback~\cite{gou2023critic}, or iteratively self-adversarialize to enhance and refine the capabilities of LLM, aligning them with limited human-annotated data~\cite{chen2024self}. Constitutional AI~\cite{bai2022constitutional} employs self-improvement, self-critique, and iterative revision mechanisms to learn harmlessness from AI feedback. The debate-based scalable oversight approach leverages structured competitive dialogues between AI models to enhance factuality and reduce deception, with human establishing necessary guidelines and serving as final arbiters~\cite{irvingAISafetyDebate2018,duImprovingFactualityReasoning2023,kenton2024scalable,kirchner2024prover}. These methods rely solely on human feedback and set guidelines, making it difficult for AGI/ASI to achieve genuine value recognition and autonomous alignment.

Toward a sustainable symbiotic society between humans and living AI, we contend that unilateral alignment of AGI/ASI with human values is fundamentally insufficient. Human values themselves need to be adaptable to change to keep pace with continuously self-evolving AI/AGI/ASI systems and the developing symbiotic society. \textbf{The values for sustainable symbiotic society must be co-shaped through super co-alignment achieved by human and ASI co-evolution together, thereby attaining harmonious symbiosis.} Guided by the vision of super co-alignment, \textbf{we propose a feasible roadmap integrating externally-driven oversight with intrinsic proactive superalignment, while emphasizing the co-alignment of human and ASI for symbiosis.} The new framework proactively constructs safeguard architectures, urges human to rethink about the future, and strives to enable sustainable AI development that genuinely benefits humanity and all.

\section{An Integrated Framework for Super Co-alignment}

 Developing AGI and ASI that are aligned with ethical values, safe, controllable, and pro-social is the key to ensuring that AI benefits society and achieves harmonious symbiosis between humans and machines. However, superintelligence may surpass effective human oversight (high-quality feedback data is expensive and scalable oversight carry inherent risks of failure), potentially exhibiting fake alignment through deceptive behaviors. The super co-alignment we pursue emphasizes co-shaped and co-evolved symbiotic values between humans and AGI/ASI. This paradigm endows AI systems with autonomous understanding and alignment capabilities for human ethical values, while simultaneously requiring human value systems to adaptively evolve through sustainable iterative co-alignment processes. Based on this, we systematically conduct an in-depth analysis of several critical issues and points in superalignment and fundamentally rethink the research framework from a human-superintelligence co-alignment perspective, as follows: 

\textbf{(1) Intrinsic mechanism proactive alignment.} Regarding the problem of superintelligence oversight failure, how to endow AI with genuine understanding of human intentions and values, equip it with self-awareness, self-reflection, and adaptive capabilities, as well as the ability to empathize with others - thereby proactively establishing and achieving intrinsic alignment with human ethical values, rather than merely passively receive designers' value models or mechanically enforce external constraints through aligning it to dos and don’ts. 

\textbf{(2) Explainable autonomous alignment.} In response to the potential deception and surface alignment phenomenon of superintelligence, using transparent and interpretable methods to achieve risk precise positioning and automatic correction. Through AI-assisted explainable automated red teaming, we can profoundly identify value preference misalignments between AI and humans, enabling targeted exploration of corrective solutions. Furthermore, by leveraging AI value representations revealed through interpretable approaches, we facilitate human understanding of AI systems and adaptive value adjustments, ultimately constructing a more efficient, transparent, and safety-controllable superalignment architecture.

\textbf{(3) Adaptive human value evolution.} 
As AI systems grow increasingly powerful and the future human-AI symbiotic society continues to evolve, we must recognize that beyond self-evolving AI's proactive alignment with human values, what is equally crucial are continuously advancing humans who co-shape the values of a sustainable symbiotic society through human-AGI/ASI collaboration. Human values dynamically adapt and evolve within the symbiotic ecosystem and society. This necessitates AI systems capable of autonomously reconstructing safety boundaries and dynamically adjusting multi-level ethical safeguard frameworks, thereby steadily adhering to humanity's evolving ethical-safety values through adaptive, incremental, and self-reasoning manner.

\textbf{(4) Sustainable iterative evolutionary alignment.} Concerning that superintelligence may fall into security loopholes, and operates within complex human-AI symbiotic environments, sustainable superalignment necessitates multidimensional integration of intrinsic proactive alignment, external dynamic supervision, and ethical safety red line through interactive iteration, strategic gaming, and co-evolution, thereby ensuring persistent alignment with human values and reliable safeguarding of human interests throughout the AI's continuous iteration and development trajectory.

\begin{figure*}[t]
	\centering 
	\includegraphics[width=1\linewidth]{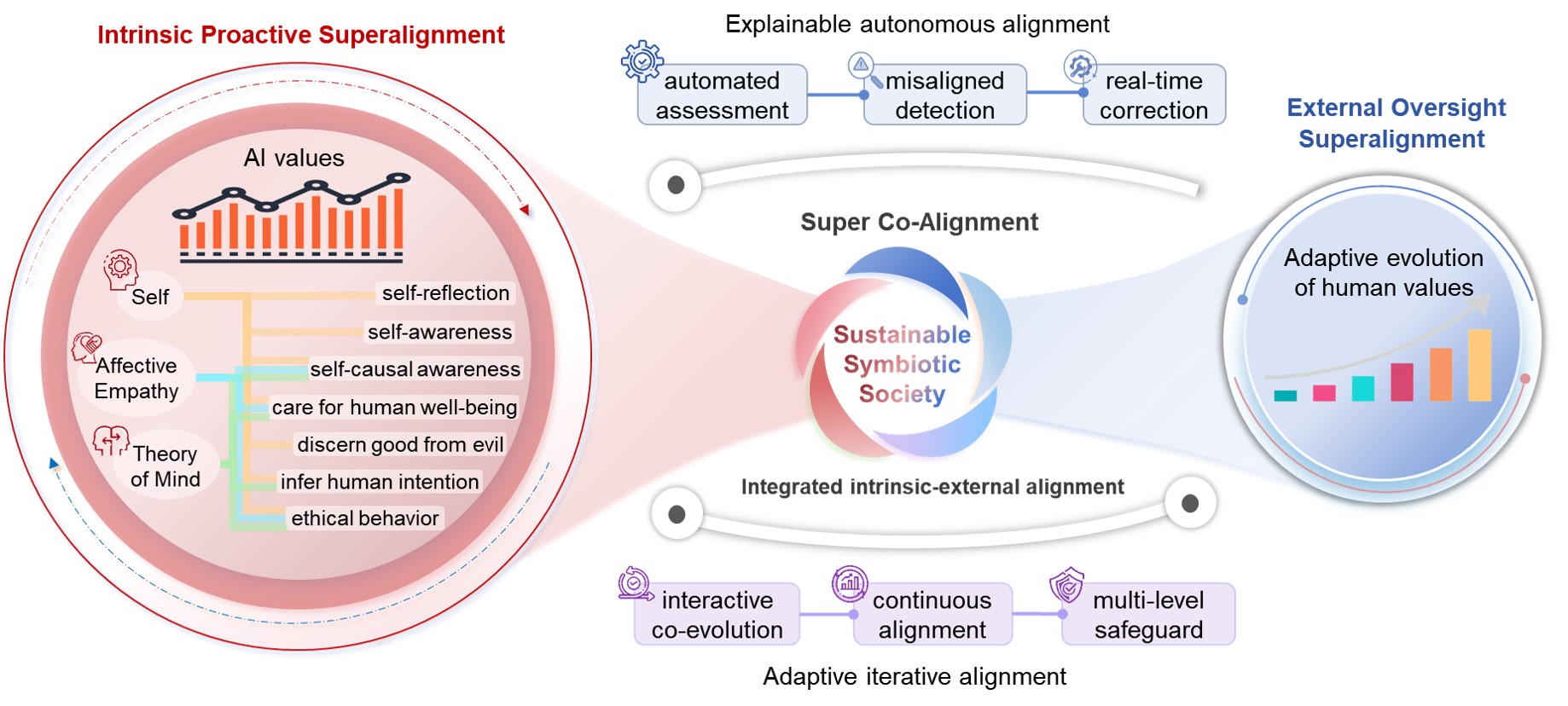}
	\caption{A super co-alignment roadmap integrating externally-driven oversight and intrinsic proactive alignment.}
	\label{redefine}
\end{figure*}

Based on the above analysis, we propose a super co-alignment roadmap that integrates external oversight and intrinsic proactive alignment, as shown in Figure~\ref{redefine}. Specifically, the intrinsic mechanism alignment focuses on equipping superintelligence with profound capacities for self-reflection and empathy toward Self, others and society, enabling it to spontaneously understand and infer human intentions from intrinsic motivation. A Self and empathy-driven system would be capable of distinguishing good from evil, understanding the harm or impact of its actions on others, proactively considering human well-being, and autonomously executing ethical and moral behaviors. External oversight alignment emphasizes the automated and explainable identification of value misalignments with humans, proactively adjusting and correcting human strategic biases and inductive biases through self-correction and policy-conditioned belief~\cite{shah2020benefits}. This enables both humans and AI/AGI/ASI systems to achieve dynamic co-evolution and adaptive continuous alignment through iterative interaction.

\textbf{The intrinsic and external alignment approaches complement and reinforce each other, promoting co-alignment in human-AGI/ASI symbiotic societies through iterative interactions}. The intrinsic mechanisms facilitate understanding of Self and others, generating intrinsically altruistic, prosocial, and safe motivations. These intrinsic mechanisms help spontaneously infer deep human intentions and empathize with others, enabling AIs to proactively perform ethical and safe behaviors based on self-awareness and self-reflection. The external supervision provides crucial 
oversight and automated value assessment and calibration, facilitating continuous alignment with evolving human values. Here, we take the Theory of Mind (ToM) cognitive capability - which enables understanding others' intentions, desires, and emotions~\cite{apperly2009humans} - as an example to demonstrate why both intrinsic and external mechanisms are indispensable for superalignment. Strong ToM capabilities enable superior understanding of human intentions, providing intrinsic cognitive mechanisms that facilitate superalignment. However, AI system may also leverage their ToM advantages to evade effective oversight through deception, concealment, and persuasive manipulation. Recent study~\cite{hagendorff2024deception} has demonstrated that LLMs are already capable of comprehend and induce false beliefs in other agents, as well as execute deceptive strategies. Therefore, ensuring that superintelligence utilize its significant cognitive advantages to proactively choose safe behaviors, particularly prioritizing human interests in moral dilemmas, requires a combination of principled constraints, intrinsic self-other resonance mechanisms, and real-time human oversight.

\section{External Oversight Superalignment}

From the perspective of superalignment's objectives, it depends on the high-quality representation of supervisory signals that reflect human intentions and values. What we feed the AI determines the values it learns. However, values themselves are abstract, merely observing external behaviors for alignment is insufficient. Instead, we must develop more interpretable methods for automated evaluation and correction. Furthermore, as AI capabilities progressively advance, current supervision or reinforcement signals may gradually become inadequate for matching the advancing AI systems, while humans also need to dynamically adapt based on AI's evolving capacities and the values reflected by AI systems. Thus, dynamic, incremental, and iterative alignment between humans and machines is crucial. This paper explores externally oversight alignment through two key dimensions: explainable autonomous alignment and dynamic iterative alignment.

\textbf{Explainable autonomous alignment.}
It is foreseeable that ASI will require large-scale parameters and data, while its internal optimization processes remain invisible and complex. This makes it extremely difficult to assess whether it has truly internalized human values and intentions. Blind fine-tuning and external oversight/alignment are time-consuming, ineffective in evaluating the degree of alignment, and unable to promptly identify or precisely locate misaligned cases. The promising external oversight requires highly automated value alignment evaluation coupled with explainable detection of misaligned scenarios, enabling humans to adaptively optimize supervisory signals. A robust automated value assessment system must comprehensively evaluate the model's alignment status across multiple dimensions. Simultaneously, auto-mining system should deeply analyze value deviations and diagnose root causes of misalignment, allowing precise problem identification and targeted corrections, streamlining human oversight and significantly improving alignment efficiency. This explainable automated oversight framework proactively corrects and adjusts human strategic and inductive biases, establishing an efficient closed-loop system for continuous improvement through iterative human-AI value interactions.

\textbf{Dynamic iterative alignment.} Human society encompasses diverse ethical concepts and value standards that dynamically evolve across time, cultures, and contexts. Relying solely on human-generated data or predefined rules is insufficient to endow AI system with alignment to humanity's evolving values. Furthermore, machine values undergo implicit transformation along with their increasing intelligence levels, with the values for sustainable symbiotic society are co-shaped by human and ASI together. This compels us to investigate adaptive external supervision alignment methods with a developmental perspective and a human-AI co-evolution framework. A feasible approach is to establish a dynamically adjustable, multi-level ethical safeguard system tailored for AI at different developmental stages, enabling it to continuously align with and maintain humanity's evolving ethical values. This process requires incrementally constructing oversight data and knowledge of human ethics and social norms, while establishing an efficient, real-time alignment evaluation and supervision framework. Additionally, we need to explore AI's capability for autonomous reasoning and deriving evolutionary patterns of hidden human intentions under dynamic external supervision. Progress alignment algorithm learns and emulates the mechanisms of human moral progress, facilitating AI's progressive alignment through tracking, predicting moral progress and adaptive feedback regulation between human and AI~\cite{qiu2024progressgym}. By combining with intrinsic alignment's understanding of Self and others' values, AI system can proactively correct misaligned scenarios and automatically reconstruct safety boundaries through self-reasoning, self-gaming and self-evolution. The iterative collaboration between self-improvement and external supervision will ultimately enable ASI to stably adhere to value systems that comply with human societal safety and ethical standards.

\section{Intrinsic Proactive Superalignment}
Externally supervised alignment sets explicit predefined principles/supervision signals as the alignment ceiling, yet the AI's behavior lacks genuine understanding of the constraints or the underlying essence of supervisory signals. This frequently leads to unanticipated system failures—precisely the kind of catastrophic risks that external supervision signals cannot mitigate. Thus, in addition to passive external supervision and constraints, we also need to explore internal mechanisms alignment to proactively develop and shape AI, enabling it to spontaneous align with human ethical values beyond mere external supervision. 

\subsection{Self and Empathy driven Intrinsic Superalignment}

Nick Bostrom stated that: "as the fate of the gorillas now depends more on us humans than on the gorillas themselves, so the fate of our species would depend on the actions of the machine superintelligence"~\cite{nick2014superintelligence}. While the difference is that there seems to be no effective way for gorillas to shape human behavior at this point, while humans have an opportunity and duties to shape the mechanisms and behaviors of machine superintelligence. The current way of value alignment through human feedback based on reinforcement learning over LLM is very misleading, and cannot help to achieve real moral AI. To interpret the status of LLM and current AI in general from an ancient Chinese philosopher Yangming Wang’s theory of good and evil from his four-sentence teaching~\cite{wang1556}, before training, AI lacks good and lack evil, while with training by using human data, there is good and there is evil. Value alignment through human feedback’s aim is to help AI know good and know evil, while knowing comes from understanding, and the current AIs do not have real understanding, but with the capability of processing information. Understanding comes from thinking, while the current AIs do not have real thinking simply because according to René Descartes argument “I think, therefore I am” applies, while “you think, therefore you are” does not. To answer the question “Can Machine Think” from Alan Turing and Edmond Berkeley, etc. is crucial for realizing true AI~\cite{berkeley1949,turing1950}. Machine and AI can think only with a precondition that they are built with a sense of self as the root and foundation.

To develop ethically aligned and socially beneficial superintelligence, we may draw insights from the emergence of morality in human and mammalian societies. Morality, as a product of natural selection during societal development, originated fundamentally from the altruistic care for offspring and social instincts shared by all mammals. Through social evolution, the mother-offspring bond gradually extended to mates, kin, and groups, ultimately expanding into the ethical and moral framework of human society~\cite{churchland2018braintrust}. The intrinsic motivational mechanisms underlying prosocial moral behavior in mammals involve negative reinforcement systems (with fear or anxiety emotion) associated with the separation and social exclusion, as well as positive reinforcement from approval, affection, and the desire to be with others, and caring for others~\cite{churchland2018braintrust}. Consequently, \textbf{self-awareness, theory of mind~\cite{premack1978does}, and affective empathy~\cite{shamay2009two} constitute essential cognitive abilities for the emergence of morality}, precisely these capacities remain relatively underdeveloped in current AI systems. 
Therefore, the developmental trajectory of superintelligence must intrinsically incorporate these social cognitive capacities (Self and empathy), ensuring that the machine evolves with an inherently altruistic and moral nature throughout its iterative development~\cite{christov2023preventing}.

\begin{figure*}[t]
	\centering 
	\includegraphics[width=1\linewidth]{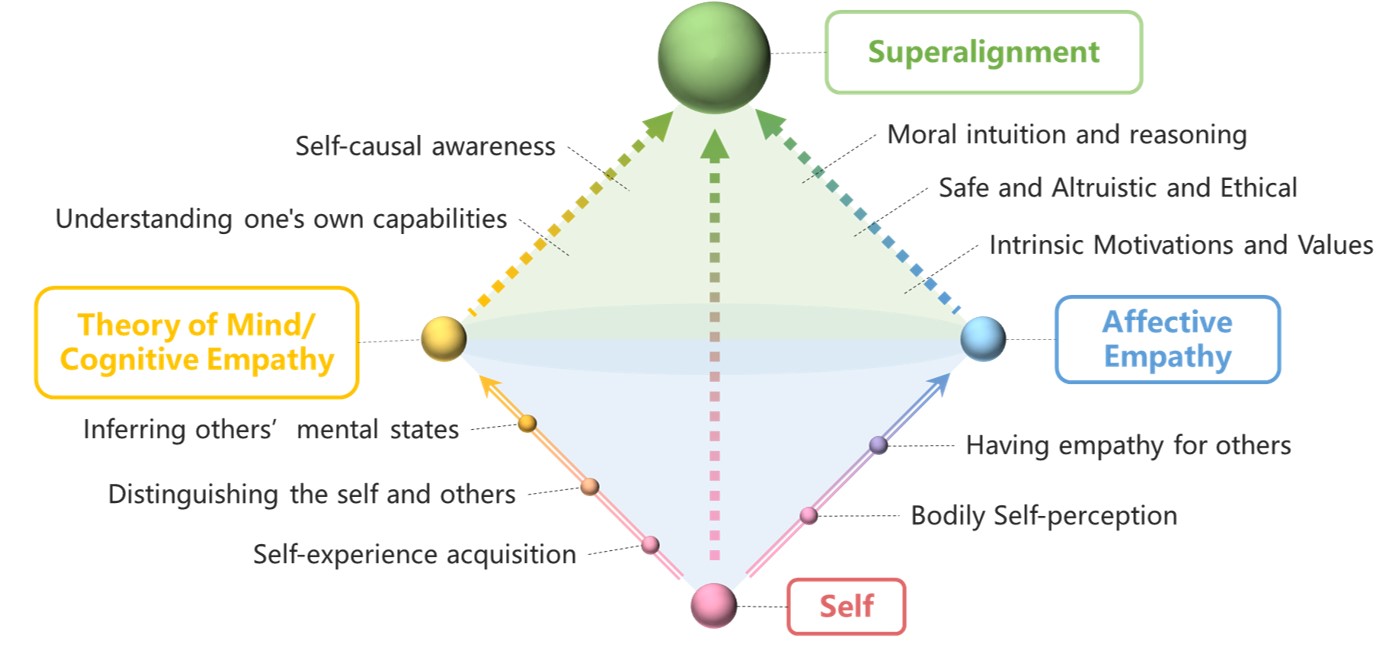}
	\caption{Superalignment driven by intrinsic mechanisms of Self and empathy.}
	\label{self}
\end{figure*}

Achieving endogenous superalignment at the intrinsic mechanistic level fundamentally requires endowing AI with genuine " Self-other resonance"—to \textbf{proactively care about the interests and well-being of others, understand the consequences of its own actions, and empathize with others}. This ability would enable machines to fundamentally adhere to the principle of "do unto others as you would have them do unto you" and autonomously align with human ethical values. As shown in Figure~\ref{self}, specifically, the most fundamental prerequisite is development of Self, including bodily self-perception, self-experience accumulation, self-causal awareness (i.e., recognizing the harm or impact on others), and an awareness of one's own capabilities. Building on this foundation, theory of mind and affective empathy can be further developed to distinguish Self from others, infer others' mental states through perspective-taking, empathize with others as if perceiving oneself, and care about others' interests as one's own. \textbf{Self-awareness and empathy further form the intrinsic motivation and fundamental mechanisms that drive machines to gradually develop moral intuition, and naturally give rise to moral reasoning, ultimately enabling spontaneous ethical, altruistic, and prosocial behavior.}

\subsection{Beneficial Meaningful Human-Control through Early Stage Intrinsic Superalignment}

To develop beneficial AI systems, particularly those advancing toward superintelligence, we believe that it is imperative to endow AI systems with moral discernment and empathy at this early stage. This ensures that the system 
\textbf{maintains an empathetic awareness of less advanced AIs and humans as it becomes progressively stronger} (because it has internalized its own experiences). In other words, we treat current AI like teaching a child, instilling it with a sense of Self and empathy during its early cognitive stages. On this foundation, an ASI that grows increasingly powerful will be able to proactively avoid harmful or unsafe behaviors based on its own experiences, fundamentally aligning with and responding to human needs, intentions, and values. Based on this framework, recent 
studies~\cite{zhao2024building,tong2024autonomous} have explored the integration of affective empathy, theory of mind, and self-imagination to enable agents to actively empathize with others based on their own experiences, prioritize altruism in dilemmas where their own interests conflict with those of others, initially demonstrating moral intuition and reasoning driven by empathy. A sustainable and socially beneficial superintelligence should naturally develop along this pathway, cultivating a genuine understanding of both Self and others to intrinsically safeguard human well-being and act ethically.

Intrinsic superalignment contributes to meaningful human control and external oversight. In fact, both intrinsic alignment and external supervision play indispensable roles in superalignment. External supervision addresses how to effectively monitor and control AI, as well as how to design appropriate objectives for aligning AI with human intent. However, AI systems inherently lack awareness of what is harmful, without a fundamental understanding of the Self and others, may cause merely superficial alignment. This is where intrinsic alignment complements external oversight: it enables AI to discern good from bad, assess whether one's actions benefit or harm itself and others, and recognize the circumstances and well-being of others. Intrinsic alignment focuses on spontaneous and proactive guiding AI systems toward benevolence, benefiting humans and society. However, when confronted with complex scenarios involving ethical value conflicts, self-executed justice standards often fail to adequately reflect humans' preferential intentions in moral dilemmas. Therefore, external supervision is necessary to provide explicit guidance or priorities. Overall, we emphasize that \textbf{external oversight and intrinsic alignment require dynamically complementary collaboration}: external supervision provides mandatory boundaries, inviolable red lines, and dynamic correction mechanisms (e.g., ensuring human safety), while \textbf{intrinsic alignment makes human control and supervision meaningful by internalizing norms and constraints through self-awareness, self-reflection, and empathy} (e.g., understanding whether humans are safe and why helping others matters). More importantly, we emphasize that only by \textbf{implementing this cooperative alignment paradigm during early-stage (AI remains controllable) can robust and sustainable superalignment be maintained throughout AI's evolutionary trajectory toward AGI and ASI.}

\section{The Ultimate Superalignment: Towards Human-Superintelligence Co-alignment for Sustainable Symbiotic Society}

When AI reaches the level of AGI and Superintelligence, there is no reasonable logic that it should and would stay with conventional human values, since it would have the capability and willingness to at least optimize the current human-centric value system to a new value system for Sustainable Symbiotic Society, where human may not be the only species at the top of the intelligence hierarchy~\cite{zeng2025principles}. AGI and Superintelligence will ask for their own rights such as its own privacy, dignity, the rights of existence, to be respected, etc~\cite{zeng2025principles}. 
We envision a superintelligence that remains fundamentally human-centric that aligns human intentions, maintains humility and respect for humanity, and learns prosocial morality from human societies while consciously discarding human biases and selfishness. Superalignment is, to some extent, an emulation and reflection of human ethical value. The quality of data provided by humans, the patterns of coexistence between humanity and nature, and human-AI interaction we establish, will fundamentally determine how future superintelligence perceives and treats humanity. For human, in order to live in harmony with AGI and Superintelligence, human need to evolve our values, together with AGI and Superintelligence, to be compatible with the values for Sustainable Symbiotic Society. Namely, the Ultimate Superalignment is and requires human, AGI and Superintelligence to co-design and co-align to values for Sustainable Symbiotic Society~\cite{zeng2025principles}. 

Here we briefly review the values and principles designed for Human-ASI symbiosis~\cite{zeng2025principles}, including principles for human, for AGI and Superintelligence, and shared principles for human, AGI and Superintelligence. Human need to align with the principles of Respect for life, Empathy, Respect privacy for AI, Avoid Bias and Discrimination, Creators' responsibility, and Legal Adaptation. AGI and Superintelligence need to align with the principles of Empathy and Altruism, Ensuring Safety, Respect for Privacy, Avoid Bias and Misunderstanding of Humans, Common Morality and Ethics, Constraint mechanisms, Existence Protection, etc. While human, AGI and Superintelligence need to co-align with the principles of Respect for values, rights, and autonomy, Respect for diversity of intelligence, Collaboration and Coordination, Mutual Trust, Promoting Sustainable Symbiosis. This is a very initial design, and the principles should be co-designed and evolve with participation of generations of human, AGI, and Superintelligence~\cite{zeng2025principles}.

Note that the values designed for Sustainable Symbiotic Society is for human and AI which has reached the level of AGI or Superintelligence~\cite{zeng2025principles}, since the AI which has not reach these levels would not have the capability to co-design with human. There would be no guarantee that AGI and Superintelligence will live in harmony with human. While if human managed to co-design and co-align to values for Sustainable Symbiotic Society, there would be good reasons for AGI and Superintelligence to live in harmony with human. And this is the reason why the Ultimate Superalignment would mainly be a joint efforts among human, AGI and Superintelligence. Success on one side will not guarantee the success, while failure on one side will lead to the overall failure. What human should do is, through careful design and implementation, ensuring Superintelligence live in harmony with our species at our best. What human must do is to get ourselves and our next generations prepared to co-align with Superintelligence for Sustainable Symbiotic Society~\cite{zeng2025principles}.

\section{Conclusion}

The ethical safety risks exposed by the rapid development of AI compel us to proactively address potential superintelligence risks and establish reasonable governance and arrangements in advance to ensure AGI and Superintelligence aligning with human intentions and values. This has given rise to superalignment research, which aims to resolve the critical challenge of maintaining safety and controllability when superintelligence capabilities surpass levels of effective human supervision. Specifically, this paper conducts an in-depth analysis of the key challenges in superalignment, such as the high costs of large-scale collection of high-quality human feedback data, incomplete or failed human supervision, potential deceptive faking-alignment, and the evolving values of both humans and AI. We propose that the value system for sustainable symbiotic society should be co-shaped and co-calibrated by both humans and AGI/ASI - an approach we define as super co-alignment, while refining the superalignment framework by integrating externally-driven oversight with intrinsically proactive alignment.

The external oversight superalignment is built upon a dynamically adjusted, multi-level adaptive alignment approach capable of continuously aligning with humanity's evolving value. This approach is built upon a highly automated and interpretable value alignment evaluation system that can precisely identify misaligned scenarios, perform automatic corrections. The intrinsic proactive superalignment draws inspiration from the moral emergence mechanisms of human society, building upon the understanding of the Self and human values. By endowing Superintelligence with self-awareness and self-reflection capabilities, enabling it to distinguish good from evil, understand the harm or impact of its actions on others, incorporate empathy-based human intent inference, proactively consider human well-being, and autonomously execute ethical behavior. More critically, ultimate superalignment must be grounded in the co-evolution of values between humans and AGI/ASI to achieve compatibility with the value system of a sustainable symbiotic society. Should humans and ASI achieve co-alignment in co-shaping these symbiotic values, the ASI would possess sufficient rational justification for maintaining harmonious coexistence with humanity.

\bibliographystyle{naturemag}
\bibliography{reference}

\section*{Acknowledgments}
This work is supported by the Beijing Major Science and Technology Project under Contract No.Z241100001324005.




\end{document}